\documentclass{article}

\usepackage{multirow}
\usepackage{microtype}
\usepackage{graphicx}
\usepackage{subfigure}
\usepackage{booktabs} 

\usepackage{hyperref}

\usepackage[accepted]{icml2024}

\usepackage{amsmath}
\usepackage{amssymb}
\usepackage{pifont}
\usepackage{mathtools}
\usepackage{amsthm}

\usepackage[capitalize,noabbrev]{cleveref}
\newcommand{\cmark}{\ding{51}}%
\newcommand{\xmark}{\ding{55}}%
\theoremstyle{plain}

\theoremstyle{definition}

\theoremstyle{remark}

\usepackage{subcaption} 

\usepackage[textsize=tiny]{todonotes}

\usepackage{color,soul}
\icmltitlerunning{COPAL: Continual Pruning in Large Language Generative Models}

\begin{document}

\twocolumn[
\icmltitle{COPAL: Continual Pruning in Large Language Generative Models}

\icmlsetsymbol{rmk}{*}

\begin{icmlauthorlist}
\icmlauthor{Srikanth Malla}{comp}
\icmlauthor{Joon Hee Choi}{comp}
\icmlauthor{Chiho Choi}{comp}
\end{icmlauthorlist}

\icmlaffiliation{comp}{Samsung Semiconductor, San Jose, USA}
\icmlcorrespondingauthor{Joon Hee Choi}{jh4.choi@samsung.com}
\icmlcorrespondingauthor{Chiho Choi}{chiho1.choi@samsung.com}

\icmlkeywords{Machine Learning, ICML}

\vskip 0.3in
]

\printAffiliationsAndNotice{ }  %

\begin{abstract}

Adapting pre-trained large language models to different domains in natural language processing requires two key considerations: high computational demands and model's inability to continual adaptation. 
To simultaneously address both issues, this paper presents COPAL  (\textbf{CO}ntinual \textbf{P}runing in \textbf{A}daptive \textbf{L}anguage settings), an algorithm developed for pruning large language generative models under a continual model adaptation setting. While avoiding resource-heavy finetuning or retraining, our pruning process is guided by the proposed sensitivity analysis. The sensitivity effectively measures model's ability to withstand perturbations introduced by the new dataset and finds model's weights that are relevant for all encountered datasets. As a result, COPAL allows seamless model adaptation to new domains while enhancing the  resource efficiency. 
Our empirical evaluation on a various size of LLMs show that COPAL outperforms baseline models, demonstrating its efficacy in efficiency and adaptability. 


\end{abstract}

\section{Introduction}


The advent of Large Language Models (LLMs) such as GPT-3~\cite{brown2020language} and LLaMA~\cite{touvron2023llama} has been a landmark in natural language processing (NLP). Adapting these pre-trained LLMs to diverse domains has offered unprecedented capabilities in various NLP tasks including language understanding and generation~\cite{gururangan2020don}. However, there are two primary challenges associated with this approach: (i) extensive computational requirements; and (ii) limited model adaptability. 
Considering the large-scale models and datasets, the retraining procedure necessitates considerable computational effort, which often limits its feasibility in resource-constrained environments. Moreover, once the models are updated for certain tasks or domains, they may not exhibit the same level of performance when confronted with data that deviate significantly from what they were trained on (known as "catastrophic forgetting").  

\begin{figure}
    \centering
    \includegraphics[width=1\linewidth]{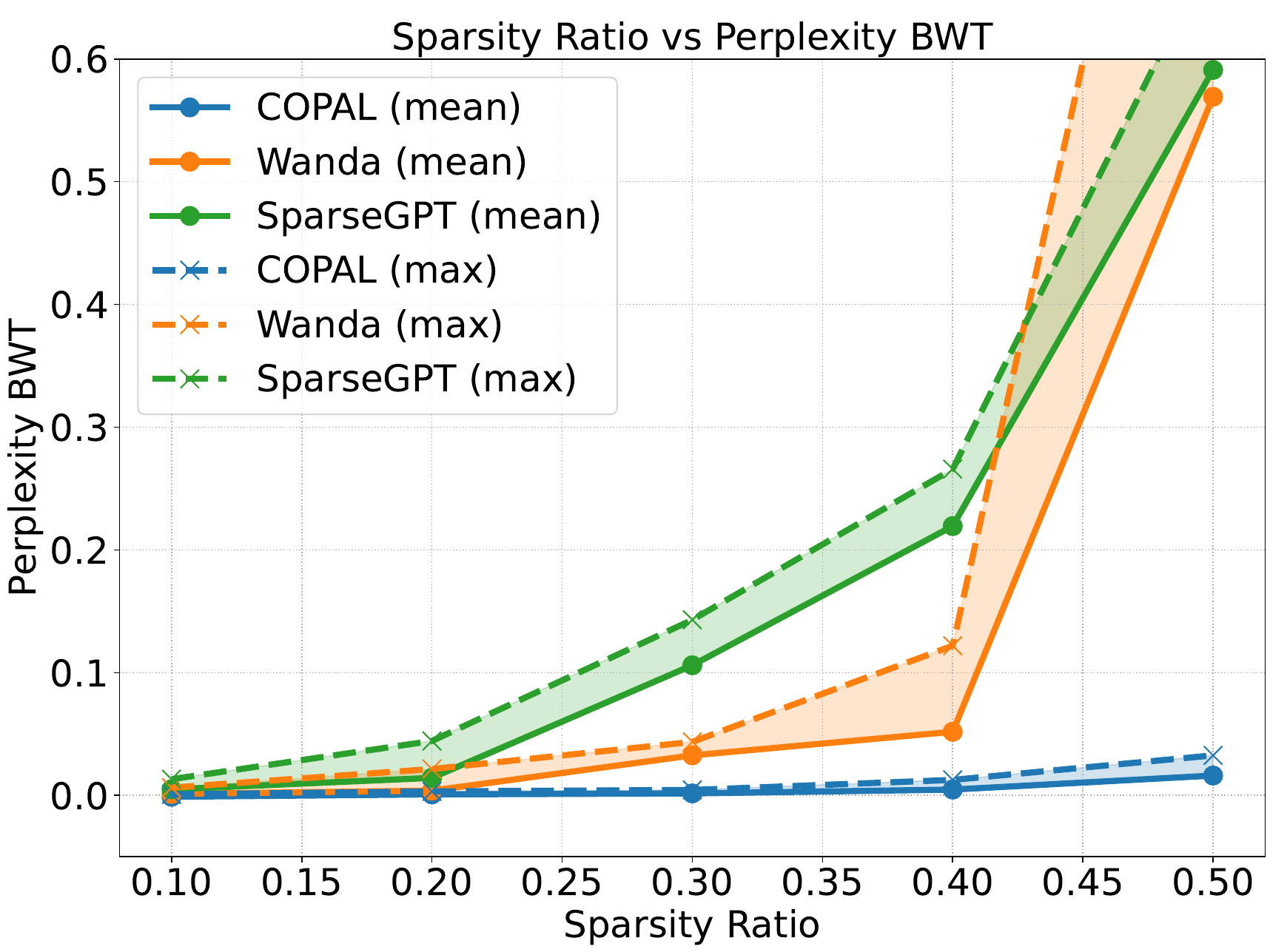}
    \caption{Motivation: average (mean) and worst (max) case scenarios of backward transfer (BWT) in perplexity with increase in sparsity ratio in unstructured continual pruning of LLaMA-7B}
    \label{fig:motivation}
\end{figure}

Traditional methods have tackled these challenges separately from different standpoints, either utilizing neural network pruning~\cite{frantar2023sparsegpt} or adopting continual learning techniques~\cite{kirkpatrick2017overcoming}. The former (\textit{i.e.}, pruning) suggests to eliminate less critical connections~\cite{frankle2018lottery} or structural elements~\cite{he2017channel} of the neural network, being beneficial in enhancing the efficiency of model inference. However, it inherently lacks a flexibility in continually adapting to new information over time, as pruning involves domain-specific training~\cite{liu2020dynamic} or calibration~\cite{frantar2023sparsegpt} procedure. On the other hand, the latter (\textit{i.e.}, continual learning) aims to adapt models to evolving data streams, accommodating new tasks and knowledge without catastrophic forgetting. With its need for finetuning on a series of datasets, this approach does not alleviate the ongoing concern of substantial computational costs for model execution. Contrary to this trend, few research efforts~\cite{dekhovich2023continual} have been recently made by applying pruning methods within a continual learning context. However, they still require re-training of the base pre-trained models on the new datasets, with a potential reduction of their generalizability originially set toward various tasks. 

Our framework, COPAL (COntinual Pruning in Adaptive Language settings), is designed to simultaneously address both issues (computational inefficiency and model inadaptability), inspired by recent studies in~\cite{sun2023simple,frantar2023sparsegpt} on post-training pruning. While optimizing LLMs without involving any further training or finetuning, we find model's weights that are relevant for all previously encountered datasets. This process involves reducing the complexity and size of the model, which naturally enables the model pruning. Additionally, our solution does not require storing past datasets or reusing them for pruning. Instead, previously pruned network with remaining parameters already contains sufficient meta information to retain the performance for past datasets. Upon transitioning to the new dataset, COPAL employs a small set of calibration data to guide the pruning process based on the sensitivity analysis we propose. Since the sensitivity measures model's robustness to perturbation caused by the new dataset, this strategy enables the model to seamlessly adapt to new information while preserving previous knowledge. To the best of our knowledge, we are the first to introduce the concept of \textit{continual pruning} that addresses pruning under a continual model adaptation setting, bypassing the requirement for model re-training. This marks a substantial advancement in the field of LLM optimization.

To illustrate the effectiveness of our approach, Figure \ref{fig:motivation} demonstrates the impact of COPAL in both average and worst-case scenarios, showcasing how continual pruning can maintain performance even with increased sparsity. This highlights COPAL's ability to adeptly navigate the balance between model complexity and performance, a crucial factor in real-world applications.

Our contributions are threefold:
\begin{itemize}
    \item We explore the inherent challenges in finetuning pre-trained LLMs and provide a strategic solution to address both computational inefficiency and limited model adaptability.
    \item We propose a mathematical formulation for the concept of continual pruning, utilizing sensitivity analysis to enable the pruning process under a continual model adaptation setting.
    \item Through empirical evaluations on large-scale language models, including LLaMA-7B, 13B, 30B, 65B, we show that COPAL outperforms baseline methods, setting a new standard in LLM optimization for both efficiency and adaptability.
\end{itemize}

\begin{figure*}[t]
    \centering
    \includegraphics[width=0.99\linewidth]{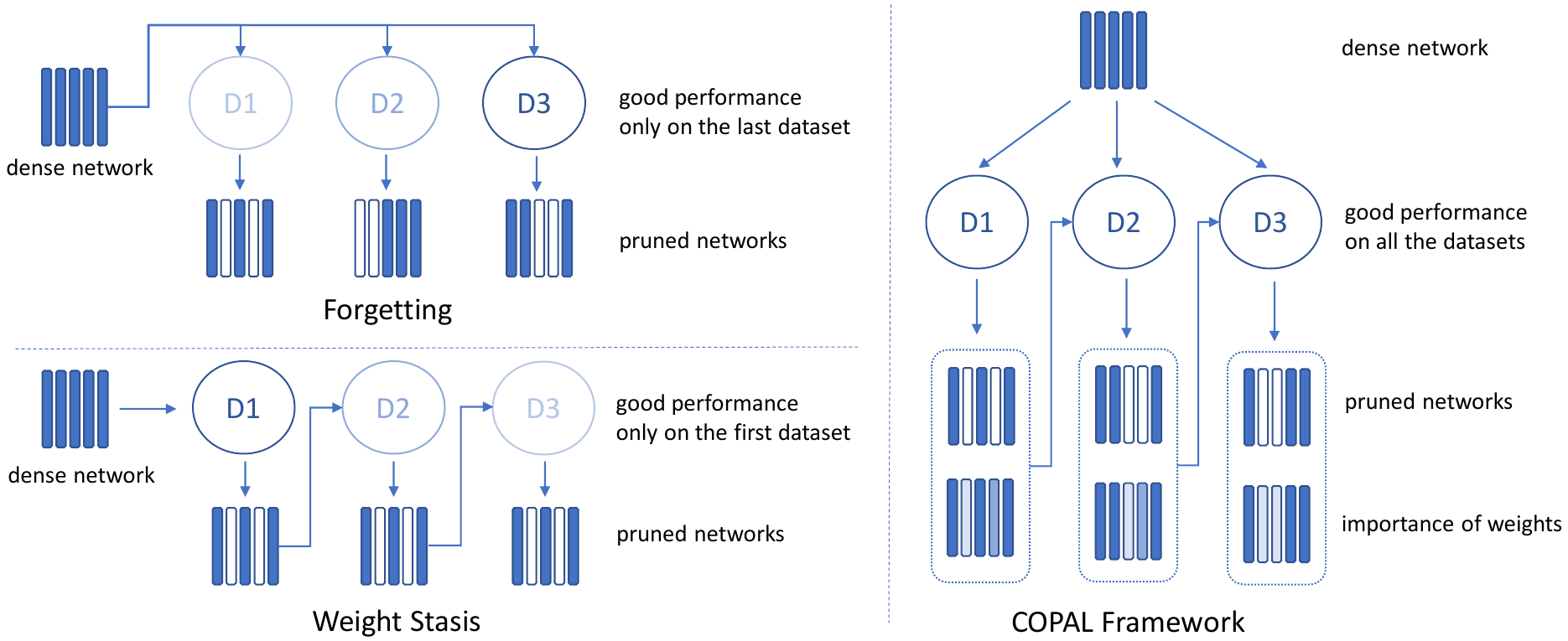}
    \caption{Overview of Forgetting, Weight Stasis and COPAL Framework in Continual Pruning. D represents calibration data used for pruning, and D1$\rightarrow$D2$\rightarrow$D3 is the dataset incremental order here.}
    \label{fig:overview}
\end{figure*}

\section{Prior Works}
\label{prior_works}
Pre-trained LLMs have shown generizability across various NLP tasks. Despite their advancements, finetuning of LLMs face challenges in computational efficiency, requiring optimization strategies like pruning for practical deployment.

\textbf{Pruning.} Pruning plays an important role in optimizing neural network architectures, especially in large models.\\
\textit{Types of Pruning:} Structured Pruning involves removing entire structural elements such as neurons, filters, or layers. Key contributions in this area include network slimming \citet{liu2017learning}, channel pruning \cite{he2017channel}, and optimizing network architectures with minimal performance trade-offs \cite{luo2017thinet, yu2018nisp}. Structured pruning is particularly relevant for simplifying large language models and enhancing their efficiency. Whereas, unstructured pruning focuses on the selective removal of individual weights. It aims to eliminate less critical connections within the network. Pioneering techniques like "Optimal Brain Damage" \cite{lecun89braindamage} and "Deep Compression" \cite{han2015deep} have significantly contributed to reducing neural network size. The "Lottery Ticket Hypothesis" \cite{frankle2018lottery} suggests the existence of smaller, effective sub-networks, which is a critical concept for large model optimization.\\
\textit{Stages of Pruning}: Pruning can be applied before training, during training, or post-training. Each stage offers unique advantages and challenges. For instance, SNIP \cite{lee2018snip} and GraSP \cite{wang2020picking} focus on identifying critical connections before training starts. On the other hand, dynamic sparse training \cite{liu2020dynamic}, soft filter pruning \cite{he2018soft} or with knowledge distillation\cite{xia2022structured, yang2022gradient} integrate pruning into the training process. Post-training techniques like those developed by \citet{sun2023simple} and \citet{frantar2023sparsegpt} are particularly relevant for large language models, enabling efficient pruning after the model has been fully trained and avoid re-training burden.


\textbf{Continual Learning.} Continual learning focuses on adapting large language models in dynamic environments,  
enabling continuous learning without losing prior knowledge.\\
\textit{Catastrophic Forgetting~\cite{mccloskey1989catastrophic}:} Methods like rehearsal techniques (e.g., Gradient Episodic Memory \cite{lopez2017gradient}, Experience Replay \cite{rolnick2019experience}), regularization methods (e.g., Elastic Weight Consolidation \cite{kirkpatrick2017overcoming}, and Synaptic Intelligence \cite{zenke2017continual}) have been developed to mitigate this issue.\\
\textit{Architectural Methods:} Adapting the network architecture for continual learning is another approach. Techniques like Progressive Neural Networks \cite{rusu2016progressive} and Dynamically Expandable Networks \cite{yoon2017lifelong} focus on evolving the model's structure to accommodate new tasks.
Continual learning, with its need for finetuning on new datasets, not only risks diminishing the base model's generalizability but also faces heightened challenges due to rising training costs of LLMs and the evolving nature of real-world NLP applications.

\textbf{Pruning in Continual Setting.}
In the field of continual learning, the role of pruning is useful for maintaining and evolving neural network structures effectively. The research by \citet{yan2022learning} on Bayesian sparse networks in continual learning settings has been impactful. Similarly, the work of \citet{dekhovich2023continual} highlights the importance of preserving efficiency in class-incremental learning. Although these studies emphasize the importance of pruning to keep neural networks efficient and adaptable to new tasks, 
the potential reduction in a base model's generalizability and increasing training costs of LLMs training are inevitable. Unlike all existing techniques, we tackle these challenges with a new method for optimizing LLMs, which bypasses the need for re-training as a notion of \textit{continual pruning}.

\section{Problem Formulation}
Continual Pruning involves the pruning of model's weights for continually evolving tasks or datasets without losing the original capability of pre-trained LLMs, such as performance on previously encountered datasets. In this section, we introduce continual pruning and explain two challenges that may arise in this new domain. 

\subsection{Continual Pruning}
Continual pruning clearly differs from pruning-enabled continual learning in its focus and methodology. While conducting pruning under the continual learning perspective, as explored in Section~\ref{prior_works}, existing methods simply adopt pruning techniques to manage network complexity during training the model in a continually evolving environment. In contrast, continual pruning specifically refers to the training-free process of pruning that occurs throughout the lifespan of the model. Continual pruning dynamically adjusts which weights are relevant for all the encountered datasets, without the need to save all the past data or reuse them for pruning. Therefore, it ensures enhanced resource efficiency with preserved model's performance over time, regardless of the task or data evolution.

\subsection{Weight Stasis in Continual Pruning}
\label{sec:wegithstasis}
Motivated by the conceptual background of calibration-guided pruning strategies (also known as post-training pruning)~\cite{sun2023simple,frantar2023sparsegpt}, continual pruning takes an advantage of their training-free process to eliminate weight parameters using a particular metric that is often derived from a calibration dataset. However, 
directly adopting their techniques is not desirable under continual model adaptation. The pruned weights already zeroed-out using their criteria (or threshold) remain consistently unchanged, which results in becoming unresponsive when transitioning the model from one dataset to another, as illustrated in Figure~\ref{fig:overview}. We call this concept as "weight stasis" (WS). In the following, we provide a mathematical insight into weight stasis observed from calibration-guided pruning strategies. 


We consider the importance of weights, \( \mathbf{W}^*_{i}=\left| \mathbf{W}_i\cdot\mathcal{R}_i \right| \) as the basis for the pruning process, where \(\mathcal{R}_i \) is scaling or ranking the weights $\mathbf{W}_i$ with some criteria. For a given dataset \( i \), mask $\mathcal{M}_{i}$, and $\mathcal{I}$ as indicator function, the pruned weight matrix \( \mathbf{W}^p_i \) is obtained as follows:
\begin{equation}
    \begin{split}
        \mathcal{M}_{i} &=\mathcal{I}({\mathbf{W}^*_{i} < \mathcal{T}_s})= \begin{cases} 
0 & \text{if } \mathbf{w}^*_{i} < \mathcal{T}_s, \mathbf{w}^*_{i} \in \mathbf{W}^*_{i}\\
1 & \text{otherwise.}
\end{cases},\\
        \mathbf{W}^p_{i} &= \mathbf{W}_i \cdot \mathcal{M}_{i}.
    \end{split}
    \label{eq:1}
\end{equation}
To achieve a specific sparsity ratio \(s\), the threshold \(\mathcal{T}_s \) value is found automatically as detailed in Appendix~\ref{supsec:automating_thresh}.

Upon transitioning from dataset \( i \) to dataset \( i+1 \), the pruning mechanism starts with \(\mathbf{W}_{i+1}=\mathbf{W}^p_i \) as the initial weight matrix. This initial matrix already has a set of pruned (zeroed-out) weights. Following Equation~\ref{eq:1}, mask $\mathcal{M}_{i+1}$ is derived as 
\begin{equation}
    \begin{split}
    \mathcal{M}_{i+1} &= \mathcal{I}\left( \mathbf{W}^*_{i+1} < \mathcal{T}_s  \right)\\
    &=\mathcal{I}(\left| \mathbf{W}_{i+1} \cdot \mathcal{R}_{i+1} \right|< \mathcal{T}_s) \\
    &=\mathcal{I}(\left| (\mathbf{W}_i \cdot \mathcal{M}_{i}) \cdot \mathcal{R}_{i+1}\right|< \mathcal{T}_s) \\
    &= \mathcal{I}(\left| \mathbf{W}_i \cdot \mathcal{R}_{i+1}  \right| \cdot \left| \mathcal{M}_{i}\right|< \mathcal{T}_s)\\
    &\begin{cases} 
    0 & \text{if } m < \mathcal{T}_s, m \in \left| \mathbf{W}_i \cdot \mathcal{R}_{i+1}  \right| \cdot \left| \mathcal{M}_{i}\right|\\
    1 & \text{otherwise.}
    \end{cases},\\
    &= \mathcal{M}_{i}, \\
    \mathbf{W}_{i+2} &= \mathbf{W}_{i+1} = \mathbf{W}_i \cdot \mathcal{M}_{i}.\\
    \end{split}
  \label{eq:2}
\end{equation}
\begin{table}[t]
    \caption{Weight Stasis and Forgetting problems on pruning techniques in a continual setting, tested with unstructured pruning with 50\% sparsity ratio on LLaMA-7B. A-BWT is average of backward transfer with all incrementa dataset permutations. S-INIT is sequential initialization of weights.}
    \label{tab:weight_stasis}
    \centering
     \scalebox{0.93}{
     \begin{sc}
    \begin{tabular}{c c c c}
\toprule
       method & s-init&problem & a-bwt \\\midrule \midrule
       magnitude & - &weight stasis& WS\\
       \midrule
       sparse gpt & \xmark &Forgetting&0.591  \\
       &\cmark &weight stasis& WS \\\midrule
       wanda &\xmark &Forgetting&  0.569  \\
       &\cmark &weight stasis& WS \\       \bottomrule
    \end{tabular}
    \end{sc}
    }
\end{table}
Since the masked weights in \( w^*_{i+1} \) are zero for the previously pruned weights (\( \mathbf{W}_i \)), these weights will again fall below the new threshold and remain pruned, regardless of the new dataset $i+1$. Therefore, \( \mathbf{W}^p_{i+1} \) is being equal to \( \mathbf{W}^p_{i} \) freezing these weights when transitioning to a new dataset. 
leading to the phenomenon of "weight stasis". 
This limits the model's ability to adapt to new datasets while adopting calibration-guided pruning strategies to this domain. This is also observed from the experimental results in Table~\ref{tab:weight_stasis}.
\subsection{Forgetting in Continual Pruning}
Easing weight stasis by non-sequential weight initialization, aforementioned calibration-based post-training pruning methods can be directly applied to continual settings. However, they pose a significant challenge known as catastrophic forgetting as depicted in Figure~\ref{fig:overview}. This phenomenon is observed when pruned model weights are updated using a new calibration dataset, which deteriorates the performance on previously encountered datasets or tasks. 

Table~\ref{tab:weight_stasis} demonstrates previously discussed challenges in continual pruning, evaluating exiting methods using LLaMA-7B with 50\% sparsity. With sequential initialization, all baselines encounter WS, which means pruned weights remain the same. In addition, higher A-BWT of these methods implies that they have a forgetting issue, where the performance of the on the previously encountered dataset is increasing when pruning on the current dataset. Note that COPAL reports A-BWT as 0.016 for the same evaluation as shown in Table~\ref{tab:pruning_all}.
\section{Methodology}
We propose a framework for continual neural network pruning based on sensitivity analysis to overcome two challenges of continual pruning: weight stasis and forgetting. The overview of these problems and our proposed COPAL framework is shown in Figure~\ref{fig:overview}.

\subsection{Theoretical Framework for Continual Sensitivity Analysis}
In a standard layer of a neural network, the output \( \mathbf{y}^i_j \) is computed as \( \mathbf{y}^i_j = f(\mathbf{x}^i_j, \mathbf{W}) \), where \( f \) is the layer's function, \( \mathbf{W} \) is the weight matrix of the base model, and \( \mathbf{x}^i_j \) is the input feature vector based on $j$-th input in dataset $i$. Given layer operations, we derive the sensitivity of \( \mathbf{y}\) to infinitesimal changes \( d\mathbf{x}^i_j \) and \( d\mathbf{W} \) in \( \mathbf{x}^i_j \) and \( \mathbf{W} \), yielding, 
\begin{equation}
d\mathbf{y}^i_j = \frac{\partial f}{\partial \mathbf{x}^i_j} d\mathbf{x}^i_j + \frac{\partial f}{\partial \mathbf{W}} d\mathbf{W}.   
\label{eq:3}
\end{equation}
The consideration of infinitesimal variations of \( d\mathbf{x}^i_j \) and \( d\mathbf{W} \) is analytically intractable. Therefore, we consider their approximate as $\Delta \mathbf{W}$ and $\Delta \mathbf{x}^i_j$, both of which are assumed to be sufficiently small. Equation~\ref{eq:3} is now converted to the measure of sensitivity terms 
\begin{equation}
\begin{split}
S_{\mathbf{W}}^{ij} &= \frac{\partial f}{\partial \mathbf{W}} \times \Delta \mathbf{W},\\
S_{\mathbf{x}}^{ij} &= \frac{\partial f}{\partial \mathbf{x}^i_j} \times \Delta \mathbf{x}^i_j.
\end{split}
\label{eq:4}
\end{equation}
$S_{\mathbf{W}}^{ij}$ 
serves as the theoretical measure of
sensitivity. Similarly, $S_{\mathbf{x}}^{ij}$ is the sensitivity measure of \( \mathbf{y}^i_j \) with respect to \( \mathbf{x} \). In practice, directly calculating these derivative is often challenging or infeasible, so we use the finite difference approximation of $\frac{\partial f}{\partial \mathbf{W}}$ and $\frac{\partial f}{\partial \mathbf{x}^i_j}$. With these, Equation~\ref{eq:4} can be revised as 
\begin{equation}
\begin{split}
S_{\mathbf{W}}^{ij} &= f(\mathbf{W} + \Delta \mathbf{W}, \mathbf{x}^i_j) - \mathbf{y}^i_j,\\
S_{\mathbf{x}}^{ij} &= f(\mathbf{W}, \mathbf{x}^i_j + \Delta \mathbf{x}^i_j) - \mathbf{y}^i_j.
\end{split}
\label{eq:5}
\end{equation}
These terms respectively form the impact of perturbations in \( \mathbf{W} \) with $\Delta \mathbf{W}$ and similarly those in \( \mathbf{x} \) with $\Delta \mathbf{x}$  on \( \mathbf{y} \), which are effectively converted from a theoretical concept to a practical metric. With the combined sensitivity measures, \( d\mathbf{y}^i_j \) is reformulated as
\begin{equation}
d\mathbf{y}^i_j = S_{\mathbf{W}}^{ij} + S_{\mathbf{x}}^{ij}.
\label{eq:6}
\end{equation}
More detailed proof is available in Appendix~\ref{supsec:theor_sensitivity}.

\subsection{Identification of Crucial Weights}



The loss function on the $j$-th input vector in dataset $i$ is defined as square of Euclidean norm of \( d\mathbf{y}^i_j \) as follows:
\begin{equation}
\mathcal{L}^i_j = \left\| d\mathbf{y}^i_j  \right\|^2_2.
\label{alg:loss_local}
\end{equation}
Employing Equation~\ref{eq:3}, $\mathcal{L}^i_j$ can be rewritten as
\begin{equation}
\mathcal{L}^i_j = \left\| \frac{\partial f}{\partial \mathbf{x}^i_j} d\mathbf{x}^i_j + \frac{\partial f}{\partial \mathbf{W}} d\mathbf{W} \right\|^2_2.
\end{equation}
Then, to determine the optimal perturbations in \( \mathbf{W} \) (weights to be pruned) that would minimize \( d\mathbf{y} \), we compute the gradient of \( \mathcal{L}^i_j \) with respect to \( d\mathbf{W} \) as follows:
\begin{equation}
    \nabla_{d\mathbf{W}} \mathcal{L}^i_j = 2 d\mathbf{y}^i_j \frac{\partial f}{\partial \mathbf{W}}.
\label{eq:grad_local}
\end{equation}
Equation~\ref{eq:grad_local} is proved in Appendix~\ref{supsec:gradientloss}.

By finding \( \nabla_{d\mathbf{W}} \mathcal{L}^i_j \), we can identify the output sensitivity's (\( d\mathbf{y}^i_j \)) loss function variation with respect to weight sensitivity (\( d\mathbf{W} \)).\footnote{To find the partial differential $\frac{\partial f}{\partial \mathbf{W}}$, for a linear layer where \( \mathbf{y}^i_j = \mathbf{W}x^i_j \), \( \frac{\partial f}{\partial \mathbf{W}} = \mathbf{x}^i_j \) and for generic purpose, it can be approximated as \( S_{\mathbf{W}}^{ij}\Delta \mathbf{W}^+ \). Please check Appendix~\ref{supsec: generic_layer_inverse}.}

We further introduce \( \nabla'_{d\mathbf{W}} \mathcal{L}^k \) to capture the sum of the absolute values of the individual gradients until dataset \( k \). 
\begin{equation}
\begin{split}
\nabla'_{d\mathbf{W}} \mathcal{L}^k &= \sum_{i=0}^k \sum_j \left|\nabla_{d\mathbf{W}} \mathcal{L}^i_j \right| \\
&= \sum_j 2 \left| d\mathbf{y}^k_j \frac{\partial f}{\partial \mathbf{W}} \right| + \sum_{i=0:k-1}\sum_j 2 \left| d\mathbf{y}^i_j \frac{\partial f}{\partial \mathbf{W}} \right|\\
&= \sum_j 2 \left| d\mathbf{y}^k_j \right| \left| \frac{\partial f}{\partial \mathbf{W}} \right| + \nabla'_{d\mathbf{W}} \mathcal{L}_{k-1}\\
&=  \nabla'_{d\mathbf{W}} \Tilde{\mathcal{L}}^{k}+\nabla'_{d\mathbf{W}} \mathcal{L}^{k-1},
\end{split}    
\label{eq:sum_abs_grad}
\end{equation}
where \(\Tilde{\mathcal{L}}^k\) is loss function for only dataset $k$. This metric is particularly insightful for understanding how sensitive the loss is to each individual sample, without regard to the direction of that sensitivity (positive or negative). By focusing on the magnitude rather than the direction, we aim to offer a more nuanced view of the model's robustness to perturbations in \( \mathbf{W} \) and \( \mathbf{x} \). 

Finally, we use magnitude of directional derivative $D$ of  \( \mathcal{L}^i_j \) along \(\mathbf{W}\) as our measure of importance weights denoted by \(\mathbf{W}_k^*\).
\begin{equation}
\begin{split}
\mathbf{W}^*_k &=  \sum_{i=0:k}\sum_j \left| D_{\mathbf{W}} \mathcal{L}^i_j \right|\\
&= \sum_{i=0:k}\sum_j\left|\mathbf{W} \cdot \nabla_{d\mathbf{W}} \mathcal{L}^i_j\right|\\
&=\left|\mathbf{W} \right|\cdot \sum_{i=0:k}\sum_j\left|\nabla_{d\mathbf{W}} \mathcal{L}^i_j\right|\\
&=\left|\mathbf{W} \right|\cdot \nabla'_{d\mathbf{W}} \mathcal{L}^k\\
&= \left| \mathbf{W} \cdot \right| (\nabla_{d\mathbf{W}} \Tilde{\mathcal{L}}^{k}+ \nabla'_{d\mathbf{W}} \mathcal{L}_{k-1}) \\
&=\sum_j\left| \mathbf{W}  \cdot \nabla_{d\mathbf{W}} \mathcal{L}^k_{j}\right|  + \mathbf{W}^*_{k-1}.
\end{split}    
\label{eq:imp_weights}
\end{equation}
This \( \mathbf{W}^*_k \) serves as a measure of the importance or sensitivity of the weights \( \mathbf{W} \) in affecting the loss function. A high  \( \mathbf{W}^*_k \)  value indicates that the loss function is highly sensitive to changes in the weights along the direction of \( \mathbf{W} \), making those weights particularly crucial for the model's performance.

\begin{algorithm}
\caption{\;\;COPAL}
\begin{algorithmic}[!t]
\label{alg:pseudocode}

\STATE {\bfseries Input:}  Weights $\mathbf{W}$, \; Sparsity ratio \(s\), 
\\ \;\;\;\;\;\;\;\;\;\; Datasets \( \mathcal{D}_1, \cdots, \mathcal{D}_k \),
\\ \;\;\;\;\;\;\;\;\;\; using $j$-th input data from dataset $i$ the input 
\\ \;\;\;\;\;\;\;\;\;\; and output feature of the layer $f$ are $\left( \mathbf{x}^i_j, \mathbf{y}^i_j \right)$
\STATE \textbf{Output}: Pruned weights \(\mathbf{W}^p_k\)
\STATE \textbf{Initialize}: \( \mathbf{W}^*_{0} = 0 \) 

\FOR{i = 1:k}
    \STATE $ S_{\mathbf{W}}^{ij} \leftarrow f(\mathbf{W} + \Delta \mathbf{W}, \mathbf{x}^i_j) - \mathbf{y}^i_j$
    \STATE $S_{\mathbf{x}}^{ij} \leftarrow f(\mathbf{W}, \mathbf{x}^i_j + \Delta \mathbf{x}^i_j) - \mathbf{y}^i_j$
    \STATE $ d\mathbf{y}^i_j \leftarrow S_{\mathbf{W}}^{ij} + S_{\mathbf{x}}^{ij}$
    \STATE $ \frac{\partial f}{\partial \mathbf{W}} \leftarrow  \begin{cases} 
    S_{\mathbf{W}}^{ij}\Delta \mathbf{W}^+, &\text{if $f$ is non-linear layer}\\
    x, &\text{if $f$ is linear layer}
    \end{cases}$
    
    \STATE $ \nabla_{d\mathbf{W}} {\mathcal{L}}^i_j \leftarrow 2 d\mathbf{y}^i_j \frac{\partial f}{\partial \mathbf{W}}$
    
    \STATE $ \mathbf{W}^*_i \leftarrow \sum_j\left| \mathbf{W}  \cdot \nabla_{d\mathbf{W}} \mathcal{L}^i_{j}\right|  + \mathbf{W}^*_{i-1}$
    
    \STATE $N \leftarrow \text{total number of elements in  } \mathbf{W}^*_i$
    
    \STATE $\mathcal{T}_s \leftarrow \text{Sorted } \mathbf{W}^*_{i} \left[ \lceil (1 - s/100) \times N \rceil \right]$
    
    \STATE $\mathcal{M}_{i} \leftarrow 
    \begin{cases} 
    0, & \text{if } w^{i} < \mathcal{T}_s, \;\; w^{i} \in \mathbf{W}^*_{i}\\
    1, & \text{otherwise}
    \end{cases}$
    
    \STATE $\mathbf{W}^p_{i} \leftarrow \mathbf{W}_i \cdot \mathcal{M}_{i}$
\ENDFOR

\STATE \textbf{Return}: \(\mathbf{W}^p_k\)
\end{algorithmic}
\end{algorithm}

\subsection{Algorithm}
We present the COPAL algorithm in Algorithm~\ref{alg:pseudocode}. For each dataset in the sequence $\mathcal{D}_i$, we calculate the sensitivity in output  $d\mathbf{y}^i_j$ using sensitivity measures $S_{\mathbf{W}}^{ij}, S_{\mathbf{x}}^{ij}$. The gradient of the loss function $\nabla_{d\mathbf{W}} \mathcal{L}^i_j$ is computed using the output sensitivity $d\mathbf{y}^i_j$ and the differential $\frac{\partial f}{\partial \mathbf{W}}$. We update the importance of weights ${W}^*_{i}$ using our sensitivity metric. A threshold $\mathcal{T}_s$ is determined based on the desired level of sparsity, and a pruning mask $\mathcal{M}_{i}$ is created to selectively prune weights that fall below this threshold. As the algorithm progresses through the dataset sequence, it continually updates the importance of weights and prunes accordingly, resulting in a model that retains only the most critical weights for its function across all datasets.
\section{Experiments}
\subsection{Experimental Setting}
\textbf{Setup.} Our method and the baseline models are implemented using the Pytorch framework~\cite{paszke2019pytorch}, and we employ the Hugging Face Transformers library~\cite{wolf2019huggingface}, for managing models and datasets. Our pruning experiments utilize a single NVIDIA A100 GPU with 80 GB of memory.

Following the methodologies in ~\cite{yao2022zeroquant, frantar2022gptq, sun2023simple, frantar2023sparsegpt}, we sequentially sparsify Transformer layers, significantly reducing memory requirements. Our experiments are conducted in a single step, without fine-tuning, similar to recent studies on post-training quantization and pruning of GPT-scale models, as seen in ~\cite{frantar2022gptq, yao2022zeroquant, dettmers2022llm, sun2023simple, frantar2023sparsegpt}. For calibration, we use 16 segments of 2048 tokens each, randomly chosen from the first shard of each dataset.

\textbf{Models, Datasets.} 
In this paper, we conducted a comprehensive series of experiments utilizing the LLaMA model family, which includes models with 7B, 13B, 30B, and 65B parameters, as recently introduced by \citet{touvron2023llama}. This family of models represents a significant advancement in language model capabilities, offering a range of scales to cater to different computational requirements and performance benchmarks.

Our experimentation focused on three of the most commonly used language datasets in the field: Wikitext-2~\cite{merity2016wikitext2}, the Penn Treebank (PTB)~\cite{marcus1993building}, and the Colossal Clean Crawled Corpus (C4)~\cite{raffel2020exploring}. Wikitext-2, introduced by \citet{merity2016wikitext2}, is renowned for its collection of high-quality, well-structured textual data, predominantly comprising Wikipedia articles. The PTB dataset is a widely-used resource for syntactic analysis \cite{marcus1993building}. Lastly, the C4 dataset, part of the T5 model training corpus, provides a broad and diverse range of internet text, essential for evaluating model performance across various linguistic contexts \cite{raffel2020exploring}. 
%

\textbf{Baselines.} 
In our comparison, we evaluate the standard magnitude pruning approach, as established by \citet{zhu2017prune}, alongside the more recent developments in post-training pruning works WANDA \cite{sun2023simple} and SparseGPT \cite{frantar2023sparsegpt}. Each of these techniques is applied in a layer-wise manner, facilitating scalability even in the context of exceptionally large models. 
Magnitude pruning, a method detailed by \citet{zhu2017prune}, effectively compresses models by removing weights with the smallest absolute values, which are deemed least impactful on the network's output. In contrast, SparseGPT, introduced by \citet{frantar2023sparsegpt}, integrates sparsity into the post-training process of transformer-based models. It often employs methods such as utilizing the Hessian matrix to identify weights that can be pruned with minimal loss in performance, effectively making the model sparse from the outset. Finally, WANDA, as proposed by \citet{sun2023simple}, does pruning by analyzing weight and activation distributions. This technique identifies and prunes network segments that minimally contribute to outputs.

In Table~\ref{tab:pruning_all}, baseline models employ global initialization due to weight stasis in sequential setups, as detailed in Section~\ref{sec:wegithstasis} and Table~\ref{tab:weight_stasis}. Global initialization also leads to better average perplexity compared to sequential initialization.


\textbf{Sparsity.}
In our evaluation, we specifically target the linear layers of large language models (LLMs), excluding the initial embedding layer and the final classification head. These linear layers constitute approximately 99\% of the total parameters in LLMs~\cite{sun2023simple}, making them the primary focus for pruning. Uniform sparsity is maintained across all linear layers for consistency. We explore three distinct sparsity types: unstructured sparsity, and semi-structured sparsities with 4:8 and 2:4 configurations~\cite{sun2023simple,frantar2023sparsegpt}. This allows for a more comprehensive comparison and understanding of the impacts of different sparsity structures on LLM performance.

\textbf{Evaluation Metrics.} In this study, we focus on the impact of dataset sequence permutations (\( \Pi \)) on language model performance in continual learning scenarios. We assess this through two primary metrics: perplexity and Backward Transfer (BWT).

Our primary metric for evaluating the efficacy of continual pruning methods in language models is perplexity, a well-established and robust metric \cite{yao2022zeroquant, frantar2022gptq, sun2023simple, frantar2023sparsegpt, dettmers2022llm}. Perplexity is calculated for each dataset permutation. We determine both the mean and maximum perplexity values across all permutations. The mean perplexity \( \mathbf{P}_\pi^{i,j} \) for each example \( e \) in a dataset $j$ after pruning on dataset $i$, with a permutation of datasets ordered $\pi$ is defined as:
\begin{equation}
\mathbf{P}_\pi^{i,j} = \frac{1}{M} \sum_{e=1}^{M} \exp\left(-\sum_{c=1}^{N_e} \log(p(\mathbf{w}_c^e|\mathbf{{w}_{1:c}^e}))\right)
\end{equation}
where \( M \) is the number of examples, \( N_e \) is the number of words in example \( e \), and \( p(\mathbf{w}_c^e|\mathbf{{w}_{1:c}^e}) \) represents the probability of the \( c^{th} \) word in example \( e \), $\mathbf{{w}_{1:c}^e}$ is the context or the past words in the example. 
The equations for the mean and maximum perplexity across permutations are:
\begin{equation}
\begin{aligned}
\mathbf{P}_{avg} &= \frac{1}{|\Pi|} \sum_{\pi \in \Pi, i,j \in D} \left( \mathbf{P}_\pi^{i,j} \right), \\
\mathbf{P}_{max} &= \max_{\pi \in \Pi, i,j \in D} \left( \mathbf{P}_\pi^{i,j} \right),
\end{aligned}
\label{eq:permutation_max_avg}
\end{equation}
where \( \mathbf{P}_\pi \) is the mean perplexity for the permutation \( \pi \) and  \( D \) is the total number of Datasets, 

In addition to perplexity, we incorporate Backward Transfer (BWT) as a metric, crucial in continual learning scenarios. BWT, which assesses a model's ability to retain knowledge from previous tasks, is a crucial metric in continual learning \cite{lopez2017gradient, kirkpatrick2017overcoming}.  

BWT for a particular permutation \( \pi \) is defined as the average decrease in performance of previously learned tasks after learning a new task in the permutation. It is calculated as:
\begin{equation}
\text{BWT}_\pi^{i,j} = (\mathbf{P}_\pi^{i,j} - \mathbf{P}_\pi^{j,j})
\end{equation}
where \( \mathbf{P}_{d, d} \) is the performance on dataset \( d \) immediately after learning it, and \( \mathbf{P}_{d, D} \) is the performance on the same task after learning all \( D \) datasets. 

For each dataset permutation, we calculate BWT and then determine both the average and maximum BWT values across all permutations, similar to Equation~\ref{eq:permutation_max_avg}.This approach ensures a comprehensive evaluation of model performance across diverse continual pruning conditions.
\begin{table*}[t!]
    \centering
    \begin{minipage}{\textwidth}
        \caption{Results of continual pruning on wikitext2, ptb, c4 datasets with all permutations. Unstructured with \(50\%\) sparsity ratio, Semi structured 2:4 and 4:8 pruning. A- indicates average of a metric, M- indicates maximum of a metric.}
        \label{tab:pruning_all}
        \centering
        \scalebox{0.87}{
        \begin{sc}
            \begin{tabular}{c c c c c c c c c c c c c c }
\toprule
        &&\multicolumn{4}{c}{LLAMA-7B} &&&\multicolumn{4}{c}{LLAMA-30B}\\\cmidrule(r){3-6}\cmidrule(r){9-12}
         && a-bwt & m-bwt&a-ppl & m-ppl &&& a-bwt & m-bwt&a-ppl & m-ppl \\\midrule
        Dense (no pruning) &&-&-&7.714&10.120&&&-&-&6.131&8.159\\\midrule
        \multicolumn{12}{c}{Unstructured}\\
        \midrule
        
       Magnitude  && WS & WS& 30.246 & 49.670&&& WS & WS& 10.958& 14.638 \\
       
       SparseGPT && 0.591 & 0.690& 10.166 & 13.253&&&0.395&0.730&7.452&9.520  \\

       Wanda  && 0.569 & 1.072& 9.991  & 13.626&&&0.132&0.192&7.261&9.231\\
       COPAL \textbf{(ours)}  && \textbf{0.016} &\textbf{0.032}& \textbf{9.728} & \textbf{12.585}&&&\textbf{0.007}&\textbf{0.025}&\textbf{7.240}&\textbf{9.081}\\\midrule
       \multicolumn{12}{c}{Semi structured 2:4}\\
       \midrule
       Magnitude  && WS & WS &  131.653 & 303.710 &&& WS & WS &13.757&19.139\\
       SparseGPT   &&4.365 & 6.391&  15.744   &21.771&&&1.436&3.015&9.592&12.657\\
       Wanda && 1.667 &3.192&16.154  & 23.266&&&0.493&1.079&9.363&12.183 \\
       COPAL \textbf{(ours)}&& \textbf{0.009} &\textbf{0.075}& \textbf{15.335} & \textbf{21.159}&&&\textbf{0.036}&\textbf{0.100}&\textbf{9.274}&\textbf{11.478}\\\midrule
        \multicolumn{12}{c}{Semi structured 4:8}\\
       \midrule
       Magnitude && WS & WS & 32.105&56.652 &&& WS & WS&12.998&16.881 \\
       SparseGPT && 1.929&3.045&11.924  &15.884&&&0.838&1.670&8.351&10.790   \\
       Wanda && 0.771  & 1.645&  11.929 & 16.631&&&0.231&0.486&8.094&10.353 \\
       COPAL \textbf{(ours)}&& \textbf{0.038 } &\textbf{ 0.075}& \textbf{11.734}  & \textbf{15.532}&&&\textbf{0.012}&\textbf{0.050}&\textbf{8.032}&\textbf{9.982} \\\bottomrule
    \end{tabular}
    \end{sc}
    }

    \end{minipage}
    
    \vspace{1em} 

    \begin{minipage}{\textwidth}
    \centering
    \scalebox{0.87}{
    \begin{sc}
        \begin{tabular}{ c c c c c c c c c c c c}
\hline
 && \multicolumn{4}{c}{LLAMA-13B}&&&\multicolumn{4}{c}{LLAMA-65B} \\\cmidrule(r){3-6}\cmidrule(r){9-12}
  && a-bwt & m-bwt & a-ppl & m-ppl&&& a-bwt & m-bwt&a-ppl & m-ppl  \\\midrule
  Dense (no pruning)&&-&-&6.990&9.081&&&-&-&6.139&8.878\\\midrule
  \multicolumn{12}{c}{Unstructured}\\\midrule
Magnitude && WS & WS & 28.935 & 41.368 &&&WS&WS&9.399&13.701\\
SparseGPT && 0.606  & 1.123 & 8.467  & 11.094&&&0.334&0.564& 7.235&9.874  \\
Wanda && 0.203  & 0.298 & 8.570 & 10.896&&&0.172&0.628&7.584&11.354 \\
COPAL \textbf{(ours)} && \textbf{0.029 } & \textbf{0.078} & \textbf{8.354 } & \textbf{10.818}&&&\textbf{0.001}&\textbf{0.208}&\textbf{6.791}&\textbf{8.839} 
\\\midrule
  \multicolumn{12}{c}{Semi-structured (2:4)}\\\midrule
  Magnitude && WS & WS & 28.702  & 44.072&&&WS&WS&10.544&14.704 \\
SparseGPT && 2.670 & 5.978 & 11.970  & 17.386&&&2.128&6.979&9.333&16.037 \\
Wanda && 0.871  & 1.698 & 13.209  & 18.920&&&0.184&0.389&9.230&12.665 \\
COPAL \textbf{(ours)} && \textbf{-0.007 } & \textbf{0.340} & \textbf{11.455 } & \textbf{17.155} && &\textbf{0.038}&\textbf{0.258}&\textbf{9.161}&\textbf{11.233}\\
\midrule
  \multicolumn{12}{c}{Semi-structured (4:8)}\\\midrule
Magnitude && WS & WS & 20.476 & 29.055&&&WS&WS&9.247&12.568 \\
SparseGPT && 1.248  & 2.905 & 9.773 & 13.456&&&0.907&2.992&8.363&13.358 \\
Wanda && 0.375  & 0.793 & 9.946 & 13.281&&&0.172&0.610&8.315&11.746 \\
COPAL \textbf{(ours)} && \textbf{-0.019 } & \textbf{0.160} & \textbf{9.402 } & \textbf{12.338} &&&\textbf{0.019}&\textbf{0.260}&\textbf{8.291}&\textbf{10.807}\\\hline

    \end{tabular}
    \end{sc}
    }
    
    \end{minipage}
\end{table*}

\begin{table*}[t!]
    \centering
    \begin{minipage}{\textwidth}
        \caption{Results of per dataset performance continual pruning on wikitext2, ptb, c4 datasets with all incremental permutations on LLaMA-7B,13B,30B,65B. Unstructured pruning with \(50\%\) sparsity ratio. Overall Standard deviations in Appendix~\ref{supsec:overall_std}
        }
        \label{tab:per_dataset}
        \centering
        \scalebox{0.76}{
        \begin{sc}
            \begin{tabular}{ c c c c c c c c c c c c }
\toprule
 &&\multicolumn{2}{c}{Wikitext2} &&\multicolumn{2}{c}{PTB}&&\multicolumn{2}{c}{C4} \\\cmidrule(r){3-4}\cmidrule(r){7-8}\cmidrule(r){11-12}
  &&  BWT &   PPL    & &&  BWT      &   PPL     &&&   BWT     &     PPL  \\
  \midrule
  \multicolumn{12}{c}{LLAMA-7B}\\
  \midrule
Dense (no pruning)&& -& 5.677 ± 0.000&&&-&10.120 ± 0.000&&&-&7.344 ± 0.000 \\
Magnitude &&  WS & 17.288 ± 0.000 &&& WS & 49.671 ± 0.000 &&& WS & 23.778 ± 0.000 \\

SparseGPT &&  0.687 ± 0.003  &7.352 ± 0.343 &&& 0.506 ± 0.083 &  13.141 ± 0.260 &&& 0.580 ± 0.146  &  10.005 ± 0.308 \\

Wanda &&  0.192 ± 0.077  &  7.208 ± 0.111 &&& 1.288 ± 0.378 &  13.286 ± 0.697&&&  0.227 ± 0.032 &   9.479 ± 0.116 \\

COPAL \textbf{(ours)}&&   \textbf{0.017 ± 0.011}  &  \textbf{7.144 ± 0.013} &&& \textbf{ 0.027 ± 0.021} &  \textbf{12.567 ± 0.024} &&& \textbf{0.005 ± 0.003}  &  \textbf{9.475 ± 0.004} \\\midrule

  \multicolumn{12}{c}{LLAMA-13B}\\
  \midrule
Dense (no pruning)&& -& 5.091 ± 0.000&&&-&9.081 ± 0.000&&&-&6.798 ± 0.000 \\
Magnitude &&  WS & 20.215 ± 0.000 &&& WS & 41.368 ± 0.000 &&& WS & 25.222 ± 0.000 \\

SparseGPT &&  0.465 ± 0.093  &  6.301 ± 0.241 &&& 1.121 ± 0.002 &   10.532 ± 0.561 &&&   0.233 ± 0.049  &   8.568 ± 0.121 \\

Wanda &&  0.165 ± 0.040  & 6.138 ± 0.087 &&&   0.285 ± 0.013&   10.663 ± 0.143&&&   0.159 ± 0.028 &    8.262 ± 0.082\\

COPAL \textbf{(ours)}&&   \textbf{ 0.052 ± 0.022}  &  \textbf{6.096 ± 0.032} &&& \textbf{  0.011 ± 0.018} &  \textbf{10.472 ± 0.020} &&& \textbf{0.023 ± 0.026}  &  \textbf{ 8.176 ± 0.019} \\\midrule
  \multicolumn{12}{c}{LLAMA-30B}\\
\midrule
Dense (no pruning)&& -& 4.101 ± 0.000&&&-&8.159 ± 0.000&&&-&6.131 ± 0.000 \\

Magnitude &&  WS & 7.542 ± 0.000 &&& WS & 14.638 ± 0.000 &&& WS & 10.696 ± 0.000 \\

SparseGPT &&  0.291 ± 0.039  & 5.483 ± 0.148 &&& 0.683 ± 0.048 &  9.131 ± 0.343 &&&0.211 ± 0.007  &  7.743 ± 0.105 \\

Wanda && 0.092 ± 0.033  &  5.238 ± 0.052 &&& 0.149 ± 0.025 &  9.132 ± 0.077&&&  0.156 ± 0.036 &   7.414 ± 0.082 \\

COPAL \textbf{(ours)}&&   \textbf{0.012 ± 0.010}  &  \textbf{5.229 ± 0.009} &&& \textbf{0.005 ± 0.005} &  \textbf{9.077 ± 0.004} &&& \textbf{0.005 ± 0.005}  &  \textbf{7.415 ± 0.004} \\\midrule
  \multicolumn{12}{c}{LLAMA-65B}\\
\midrule
Dense (no pruning)&& -& 3.562 ± 0.000&&&-&8.878 ± 0.000&&&-&5.978 ± 0.000 \\
Magnitude &&  WS & 5.901 ± 0.000 &&& WS & 13.701 ± 0.000 &&& WS & 8.595 ± 0.000 \\

SparseGPT &&  0.263 ± 0.047  &4.791 ± 0.136 &&& 0.561 ± 0.003 &  9.591 ± 0.281 &&& 0.178 ± 0.165  &  7.323 ± 0.147 \\

Wanda && 0.075 ± 0.003  &  4.616 ± 0.038 &&& 0.440 ± 0.188 &  10.946 ± 0.257&&&  0.001 ± 0.038 &   7.189 ± 0.027 \\

COPAL \textbf{(ours)}&&   \textbf{-0.004 ± 0.026}  &  \textbf{4.639 ± 0.018} &&&\textbf{0.066 ± 0.093} &  \textbf{8.696 ± 0.065} &&& \textbf{-0.059 ± 0.035}  &  \textbf{7.036 ± 0.033} \\\bottomrule
    \end{tabular}
    \end{sc}
    }
    
    \end{minipage}
\end{table*}


\subsection{Results}
The performance of the dense (no pruning) models sets a fundamental baseline, indicating the optimal performance without any compromise due to pruning. In the LLaMA-7B model, the dense configuration yields an average PPL of 7.714. This benchmark is slightly lower in the LLaMA-30B and LLaMA-65B models, with average PPLs of 6.131 and 6.139, respectively. These variations suggest a nuanced impact of model scale on language processing capabilities, with larger models inherently capable of better performance pre-pruning.

\textbf{Unstructured Continual Pruning.}
In Table~\ref{tab:pruning_all}, the LLaMA-7B model under unstructured pruning, COPAL exhibits a pronounced superiority. It drastically lowers average BWT to 0.016, compared to best performing baseline Wanda's 0.569, demonstrating a 97.2\% improvement. This indicates a highly effective pruning process with minimal knowledge loss. In Table~\ref{tab:per_dataset}, COPAL's PPL performance on individual datasets like Wikitext2 and PTB shows PPL performance improvement of 7.144 (compared to best performing baseline Wanda's 7.208) and 12.567 (compared to best performing baseline SparseGPT's 13.141), respectively at 50\% sparsity ratio and still shows comparable performance to the base model. For the LLaMA-13B model, COPAL reduces average BWT to 0.029 and average PPL to 8.354, indicating substantial improvements in efficiency and comprehension. This consistency across diverse datasets underlines the method's robustness and adaptability.

Shifting to the LLaMA-30B model with unstructured pruning, COPAL again leads in performance metrics. It achieves a notable 98.2\% improvement in average BWT over SparseGPT and a minimal PPL performance drop, for instance, from 8.159 (Dense) to 9.077 on PTB. These results highlight COPAL's scalability and its effectiveness in handling moderately larger models without compromising on the pruning efficiency or language understanding.

In the LLaMA-65B model, COPAL's efficacy is further amplified. The method achieves nearly negligible average BWT (0.001), a 99.7\% improvement over SparseGPT. On the C4 dataset, COPAL significantly achieves PPL performance improvement from 8.878 (Dense) to 8.696, showcasing its exceptional capability in large-scale models. This performance indicates an continual pruning strategy that not only reduces model size but also preserves, and in some aspects, enhances model comprehension abilities.

\textbf{Semi-structured N:M Continual Pruning.} In the realm of semi-structured N:M continual pruning, the COPAL technique stands out for its effectiveness across various pruning configurations, notably within the context of the LLaMA-65B model. When applied to a 2:4 pruning pattern, COPAL achieves a 98.2\% improvement in average Backward Weight Transfer (BWT), reducing it to 0.038 compared to SparseGPT's 2.128. Additionally, it enhances the average Perplexity (PPL) by 1.8\%, bringing it down to 9.161 from SparseGPT's 9.333. The proficiency of COPAL is further evidenced in the 4:8 pruning pattern, where it markedly reduces average BWT by 99.3\% to 0.019, compared to SparseGPT's 0.907, and achieves a 0.9\% improvement in average PPL, lowering it to 8.291 from 8.363. In the 2:4 pruning configuration for LLaMA-13B, COPAL achieves BWT of -0.007 and PPL of 11.455, illustrating a marked enhancement in the model's pruning efficacy. These outcomes underscore COPAL's robust adaptability and efficiency in managing semi-structured pruning, particularly within large-scale models like LLaMA-65B, where balancing efficiency and performance is paramount.

Turning attention to the LLaMA-7B and LLaMA-30B models, COPAL's superior performance persists, showcasing a consistent pattern of effectiveness in both 2:4 and 4:8 pruning patterns. For the LLaMA-7B model, COPAL not only significantly lowers the average BWT to 0.009, marking a 99.4\% improvement over next best baseline Wanda, but also decreases average PPL to 15.335, illustrating a 5.1\% improvement. In the 4:8 pruning pattern, it continues to demonstrate efficacy by reducing average BWT by 95.0\% and improving average PPL by 1.6\%, compared to SparseGPT. The LLaMA-30B model similarly benefits from COPAL's application, yielding substantial improvements in BWT and PPL across both pruning patterns, further reinforcing COPAL's capability to effectively navigate the trade-offs between both the metrics.
\begin{figure}[h!]
    \centering
    \includegraphics[width=1\linewidth]{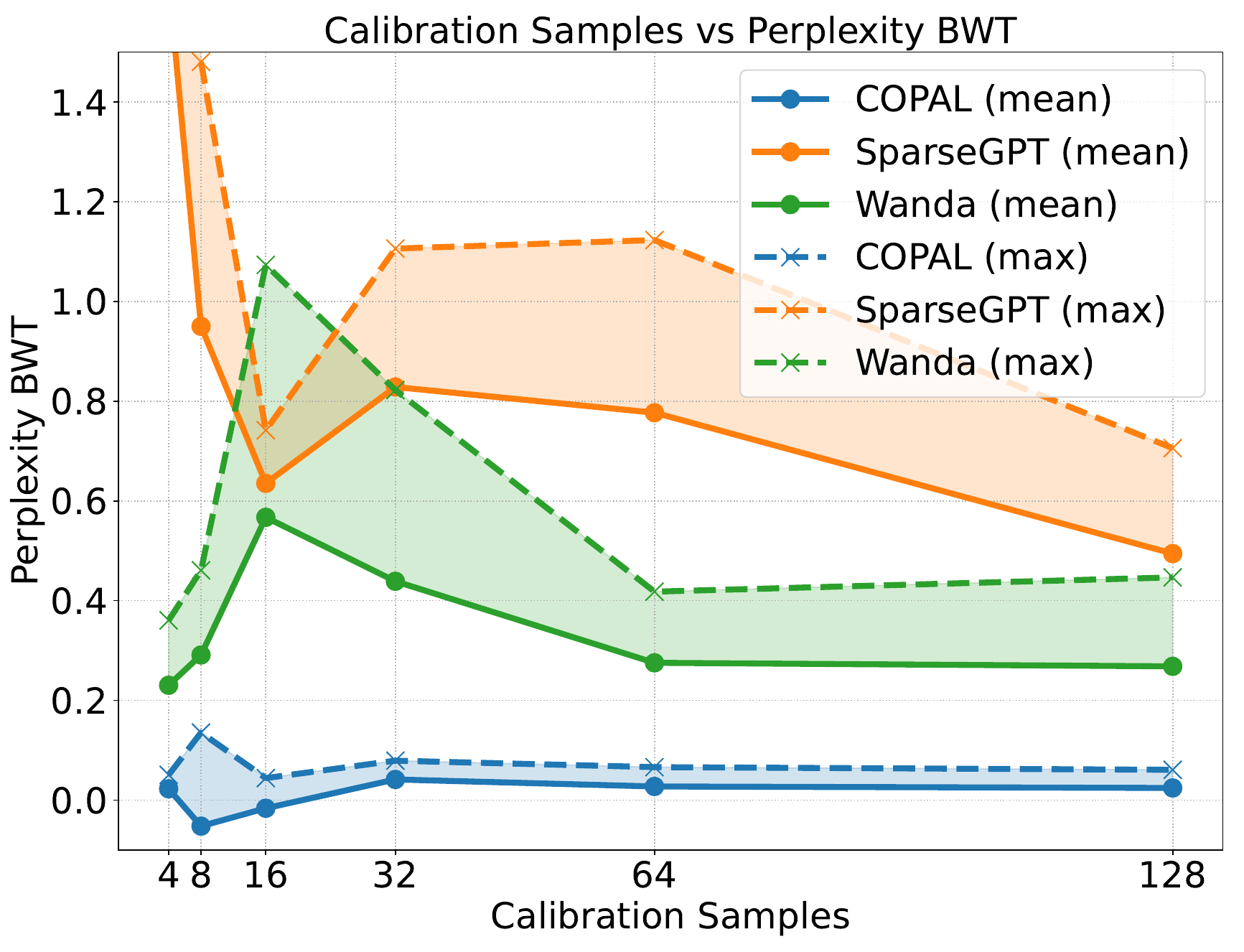}
    \caption{Motivation: average (mean) and worst (max) case scenarios of backward transfer in perplexity with increase in samples in unstructured continual pruning of LLaMA-7B with 50\% sparsity}
    \label{fig:sample_ablation}
    \vspace{-0.2cm}
\end{figure}
\subsection{Discussions}
\textbf{Ablation of Sparsity Ratio.} As shown in Figure~\ref{fig:motivation}, with increased sparsity ratios the effect of forgetting or poor performance on previous tasks is observed with other pruning methods. It demonstrates that our approach is better formulated to reduce the forgetting or maintain performance on previously observed tasks in continual pruning setting.

\textbf{Ablation of Different Sample Sizes.}
In post-training pruning usually contains a calibration data that contains a set of calibration samples. As shown in Figure~\ref{fig:sample_ablation}, COPAL is sample efficient and the average and worst case are minimal and the difference between them is also minimal. We observe that with as little as 16 samples our approach is able to perform better and stable performance even with more samples.

\section{Conclusion}
In this paper, we introduced the concept of \textit{continual pruning} that addresses pruning under a continual model adaptation setting, bypassing the requirement for model re-training. Two common problems observed in continual pruning, weight stasis and forgetting, were addressed through our proposed COPAL framework. We presented mathematical background, formulations, and full derivations of COPAL. 
Throughout the comprehensive evaluation, COPAL significantly outperformed existing methods in both Backward Transfer (BWT) reduction and Perplexity (PPL) performance, showcasing remarkable consistency and adaptability across various pruning structures, datasets and models, including LLaMA-7B, 13B, 30B, 65B.

\section*{Impact Statement}
Firstly, it's crucial to be aware of the potential for democratizing AI technology. COPAL's ability to reduce computational demands means that powerful LLMs can become more accessible to a wider range of users and organizations, potentially driving innovation across various sectors. However, this democratization also brings with it the responsibility to ensure that the models remain fair and unbiased, particularly as they adapt to changing data streams.

We must also consider the ethical implications of dynamic pruning. The adaptability of these models might lead to challenges in maintaining transparency and explainability, which are critical for user trust and accountability. As we move forward with this technology, it's essential to balance the benefits of efficiency and adaptability with the ongoing commitment to ethical AI practices and responsible usage.
\bibliography{references}
\bibliographystyle{icml2024}

\newpage
\appendix
\onecolumn

\section{Theoretical Sensitivity to Practical Metrics}
\label{supsec:theor_sensitivity}
In a general setting, let us consider a function \( \mathbf{y}^i_j = f(\mathbf{x}^i_j, \mathbf{W}) \) representing the output based on input \( \mathbf{x}^i_j \) and parameter \( \mathbf{W} \). The theoretical sensitivity \( \mathbf{S}^{ij}_{\mathbf{W}} \) is described by the derivative:
\begin{equation}
\mathbf{S}^{ij}_{\mathbf{W}} =  \frac{\partial f}{\partial \mathbf{W}} \times  \Delta \mathbf{W}
\end{equation}
This derivative represents the instantaneous rate of change of \( f \) with respect to \( \mathbf{W} \) and serves as the theoretical measure of sensitivity.

Directly calculating this derivative is often challenging or infeasible, especially when \( f(\mathbf{x}^i_j, \mathbf{W}) \) is complex. Therefore, a common approach is to approximate this derivative using finite differences. The finite difference approximation of \( \frac{\partial f}{\partial \mathbf{W}} \) is given by:
\begin{equation}
\frac{\partial f}{\partial \mathbf{W}} \approx \frac{f(\mathbf{x}^i_j, \mathbf{W} + \Delta \mathbf{W}) - f(\mathbf{x}^i_j, \mathbf{W})}{\Delta \mathbf{W}}
\end{equation}

Using this approximation, a practical sensitivity metric \( \mathbf{S}^{ij}_{\mathbf{w}} \) can be introduced as follows:
\begin{equation}
\begin{split}
\mathbf{S}^{ij}_{\mathbf{w}} &= \frac{f(\mathbf{x}^i_j, \mathbf{W} + \Delta \mathbf{W}) - f(\mathbf{x}^i_j, \mathbf{W})}{\Delta \mathbf{W}} \times \Delta \mathbf{W}\\
&=  f(\mathbf{x}^i_j, \mathbf{W} + \Delta \mathbf{W}) - f(\mathbf{x}^i_j, \mathbf{W})\\
&=  f(\mathbf{x}^i_j, \mathbf{W} + \Delta \mathbf{W}) - \mathbf{y}^i_j
\end{split}
\end{equation}
This \( \mathbf{S}^{ij}_{\mathbf{w}} \) serves as a computable stand-in for the theoretical \( \frac{\partial f}{\partial \mathbf{W}}  \Delta \mathbf{W}\). It provides a way to empirically evaluate the impact of a finite change \( \Delta \mathbf{W} \) in the parameter \( \mathbf{W} \) on the output \( f \).

By using \( \mathbf{S}^{ij}_{\mathbf{w}} \), we effectively transition from a theoretical concept to a practical, computable metric that can be used in various applications, including but not limited to, network pruning, feature selection, and model interpretation.

Similarly,
\begin{equation}
\mathbf{S}^{ij}_{\mathbf{x}} =  f(\mathbf{x}^i_j+\Delta \mathbf{x}^i_j, \mathbf{W} ) - \mathbf{y}^i_j
\end{equation}
\section{Loss Function to its Gradient}
\label{supsec:gradientloss}

The local loss function \( \mathcal{L}^i_j \) in its expanded form is:
\begin{equation}
\mathcal{L}^i_j = \left( \frac{\partial f}{\partial \mathbf{x}^i_j} d\mathbf{x}^i_j + \frac{\partial f}{\partial \mathbf{W}} d\mathbf{W} \right)^T \left( \frac{\partial f}{\partial \mathbf{x}^i_j} d\mathbf{x}^i_j + \frac{\partial f}{\partial \mathbf{W}} d\mathbf{W} \right)
\end{equation}
We aim to find \( \nabla_{d\mathbf{W}} \mathcal{L}^i_j \), which is the gradient of \( \mathcal{L}^i_j \) with respect to \( d\mathbf{W} \).

First, let's define the term inside the squared norm:
\begin{equation}
\mathbf{u} = \frac{\partial f}{\partial \mathbf{x}^i_j} d\mathbf{x}^i_j + \frac{\partial f}{\partial \mathbf{W}} d\mathbf{W}
\end{equation}

Differentiating \( \mathcal{L}^i_j \) with respect to \( d\mathbf{W} \), and treating \( \mathbf{u} \) as a function of \( d\mathbf{W} \) and \( d\mathbf{x}^i_j \), we have:
\begin{equation}
\frac{\partial \mathcal{L}^i_j}{\partial d\mathbf{W}} = 2 \mathbf{u} \frac{\partial \mathbf{u}}{\partial d\mathbf{W}}
\end{equation}

First, recall the term \(\mathbf{u}\) which is defined as:
\begin{equation}
\mathbf{u} = \frac{\partial f}{\partial \mathbf{x}^i_j} d\mathbf{x}^i_j + \frac{\partial f}{\partial \mathbf{W}} d\mathbf{W}
\end{equation}

Consider each term in \( \mathbf{u} \):

1. The term \(\frac{\partial f}{\partial \mathbf{x}^i_j} d\mathbf{x}^i_j\) does not contain \( d\mathbf{W} \), so its derivative with respect to \( d\mathbf{W} \) is zero.

2. The term \(\frac{\partial f}{\partial \mathbf{W}} d\mathbf{W}\) is directly dependent on \( d\mathbf{W} \).

So when we differentiate, we have:
\begin{equation}
\frac{\partial \mathbf{u}}{\partial d \mathbf{W}} = 0 + \frac{\partial (\frac{\partial f}{\partial \mathbf{W}} d\mathbf{W})}{\partial d \mathbf{W}}
\end{equation}
Simplifying, we get:
\begin{equation}
\begin{split}
\frac{\partial \mathbf{u}}{\partial d \mathbf{W}} &= \frac{\partial (\frac{\partial f}{\partial \mathbf{W}} \mathcal{K})}{\partial \mathcal{K}}; \mathcal{K}=d\mathbf{W}\\
\frac{\partial \mathbf{u}}{\partial d \mathbf{W}}&=\frac{\partial f}{\partial \mathbf{W}} 
\end{split}
\end{equation}
Substituting back \( \mathbf{u} \):
\begin{equation}
\frac{\partial \mathcal{L}^i_j}{\partial d\mathbf{W}} = 2 \left( \frac{\partial f}{\partial \mathbf{x}^i_j} d\mathbf{x}^i_j + \frac{\partial f}{\partial \mathbf{W}} d\mathbf{W} \right) \frac{\partial f}{\partial \mathbf{W}}
\end{equation}

Since \( \mathbf{u} = d\mathbf{y} \) for simplification, we have:
\begin{equation}
\nabla_{d\mathbf{W}} \mathcal{L}^i_j = 2 d\mathbf{y}^i_j \frac{\partial f}{\partial \mathbf{W}}
\end{equation}


To elaborate, if \( d\mathbf{W} \) is a small change in \( \mathbf{W} \), it can be thought of as \( \Delta \mathbf{W} \), a small increment. In this framework, the derivative of this small change \( \Delta \mathbf{W} \) with respect to \( \mathbf{W} \) is simply 1, or the identity matrix if \( \mathbf{W} \) is a vector or matrix.

\section{Approximating the Gradient of \( f \) with Respect to \( \mathbf{W} \)}
\label{supsec: generic_layer_inverse}

Consider a function \(\mathbf{y}^i_j = f(\mathbf{W}, \mathbf{x}^i_j) \) where \( \mathbf{W} \) is a matrix of shape \( [M, D] \) and \( \mathbf{x}^i_j \) is a matrix of shape \( [1, D] \). We wish to approximate the gradient of \( f \) with respect to \( \mathbf{W} \), denoted as \( \frac{\partial f}{\partial \mathbf{W}} \).

We introduce a small perturbation \( \Delta \mathbf{W} \) of shape \( [M, D] \) to \( \mathbf{W} \) and compute the change in \( f \) as \( S_{\mathbf{W}}^{ij} = f(\mathbf{W} + \Delta \mathbf{W}, \mathbf{x}^i_j) - f(\mathbf{W},  \mathbf{x}^i_j) \).

The approximation of \( \frac{\partial f}{\partial \mathbf{W}} \) can be expressed as:
\begin{equation}
\frac{\partial f}{\partial \mathbf{W}} \approx S_{\mathbf{W}}^{ij} \Delta \mathbf{W}^+
\end{equation}
\subsection{Full-rank of \( \Delta \mathbf{W} \)}
To ensure that \( \Delta \mathbf{W} \) is invertible, it must be a full-rank matrix. A matrix is said to have full rank if its rank is equal to the minimum of its number of rows and columns. In our case, \( \Delta \mathbf{W} \) should have a rank of \( \min(M, D) \).

\subsection{Moore-Penrose Pseudoinverse}
In cases where \( \Delta \mathbf{W} \) is not of full rank or is not square, we resort to using the Moore-Penrose pseudoinverse, denoted by \( \Delta \mathbf{W}^+ \). The pseudoinverse provides a least-squares approximation solution to the problem, and it is computed using singular value decomposition (SVD).
\section{Automating threshold for r\% sparsity}
\label{supsec:automating_thresh}
\begin{equation}
\begin{split}
\text{Sorted } \mathbf{W}^*_{i} &= \text{sort}(\mathbf{W}^*_{i})\\
 \mathcal{T}_s &= \text{Sorted } \mathbf{W}^*_{i} \left[ \lceil (1 - s/100) \times N \rceil \right]
\end{split}
\label{eq:split}
\end{equation}
To achieve a sparsity ratio of \( s\% \), the threshold is set by sorting \(\mathbf{W}^*_{i} \) and selecting the value at the \( (1 - s/100) \)-th percentile, where \( N \) is the total number of elements in \( \mathbf{W}^*_{i}\).
\section{Overall standard deviations}
\label{supsec:overall_std}
\begin{table}[h]
        \caption{Results of overall mean and standard deviation for continual pruning on wikitext2, ptb, c4 datasets with all incremental permutations on LLaMA-7B, 13B, 30B, 65B. Unstructured pruning with \(50\%\) sparsity ratio.}
        \label{sup:per_dataset}
        \centering
        \scalebox{0.85}{
        \begin{sc}
            \begin{tabular}{ c c c c }
\hline
 &&\multicolumn{2}{c}{Overall}\\\cmidrule(r){3-4}
  && BWT   &  PPL\\
  \hline
  \multicolumn{4}{c}{LLAMA-7B}\\
  \hline
Dense (no pruning)&&-&7.714 ± 1.833 \\
Magnitude& &WS & 30.246 ± 13.987\\

SparseGPT && 0.591 ± 0.122  &  10.166 ± 2.386\\

Wanda && 0.569 ± 0.5556 &   9.991 ± 2.541\\

COPAL \textbf{(ours)}&& \textbf{0.016 ± 0.017}&    \textbf{9.728 ± 2.221}\\\hline
  \multicolumn{4}{c}{LLAMA-13B}\\
  \hline
Dense (no pruning)&&-&6.990±1.634\\
Magnitude&& WS &  28.935 ± 9.026\\

SparseGPT && 0.606 ± 0.381  &  8.467 ± 1.766\\

Wanda && 0.203 ± 0.065 &    8.570 ± 1.893\\

COPAL \textbf{(ours)}&& \textbf{0.029 ± 0.028}&    \textbf{8.354 ± 1.851}\\\hline
  \multicolumn{4}{c}{LLAMA-30B}\\
\hline
Dense (no pruning)&&-&6.131 ± 1.657 \\

Magnitude && WS & 10.958 ± 2.903\\

SparseGPT && 0.395 ± 0.209  &  7.452 ± 1.520\\

Wanda && 0.132 ± 0.043 &   7.261 ± 1.595\\

COPAL \textbf{(ours)}&& \textbf{0.007 ± 0.008}&    \textbf{7.240 ± 1.576}\\\hline
  \multicolumn{4}{c}{LLAMA-65B}\\
\hline
Dense (no pruning)&&-&6.139 ± 2.173 \\
Magnitude && WS & 9.399 ± 3.235\\

SparseGPT && 0.334 ± 0.192  & 7.235 ± 1.971\\

Wanda && 0.172 ± 0.221 &   7.584 ± 2.603\\

COPAL \textbf{(ours)}&& \textbf{0.001 ± 0.078}&    \textbf{6.791 ± 1.666}\\\hline
    \end{tabular}
    \end{sc}
    }
\end{table}

\end{document}